\documentclass[runningheads]{llncs}

\usepackage[T1]{fontenc}
\usepackage{graphicx}
\usepackage{booktabs}
\usepackage{amsmath}
\usepackage{multirow}
\usepackage{algorithm}
\floatname{algorithm}{Listing} 

\usepackage{hyperref}
\usepackage{color}

\begin{document}

\title{Region-Weighted Losses and Model Fusion for Cross-Modal PET Attenuation Correction}
\titlerunning{Region-Weighted Losses and Model Fusion for PET AC}

\author{Khoa Tuan Nguyen\inst{1,2} \and
Joris Vankerschaver\inst{2,3} \and
Wesley De Neve\inst{1,2}}

\authorrunning{Khoa et al.}

\institute{IDLab, ELIS, Ghent University, Belgium \and
Center for Biosystems and Biotech Data Science, Ghent University Global Campus, Korea \and
Dept.\ of Mathematics, Computer Science and Statistics, Ghent University, Belgium\\
\email{\{khoatuan.nguyen,joris.vankerschaver,wesley.deneve\}@ghent.ac.kr}}

\maketitle

\begin{abstract}
We describe our approach to the Big Cross-Modal Attenuation Correction (BIC-MAC) challenge, which asks for a pseudo-CT in Hounsfield Units to be synthesized from Non-Attenuation-Corrected PET (NAC-PET), DIXON MRI and a topogram, and scores both the pseudo-CT and the Attenuation-Corrected PET (AC-PET) reconstructed from it.
Three ideas carried our improvements over the organizers' 3D U-Net baseline.
The loss matters more than the architecture: we compute the $L_1$ error in the Carney attenuation-coefficient ($\mu$) space that the CT metric itself uses, weighted by anatomical region.
Only once that loss was in place did the unregistered DIXON MRI work as extra input channels.
A fixed convex combination of two independently trained models then beat both of its members on three of the four metrics and ranks first overall on the public validation leaderboard.

\end{abstract}

\section{Task and Evaluation}

Given a subject's NAC-PET, two-phase DIXON MRI and a 2D topogram, we must predict a whole-body pseudo-CT in Hounsfield Units (HU), which the organizers pass through a fixed reconstruction pipeline built on STIR~\cite{thielemans2012stir}.
Four metrics are reported, all lower-is-better: the whole-body CT $\mu$-map Mean Absolute Error (MAE), computed after converting HU to linear attenuation coefficients with the Carney bilinear model~\cite{carney2006hu2mu}; the whole-body Standardized Uptake Value (SUV) MAE; an organ bias; and a dataset-level brain-outlier score.
The final ranking is the mean of the per-metric ranks rather than the mean of the metrics themselves.

\begin{figure}[t!]
\centering
\includegraphics[width=0.85\textwidth]{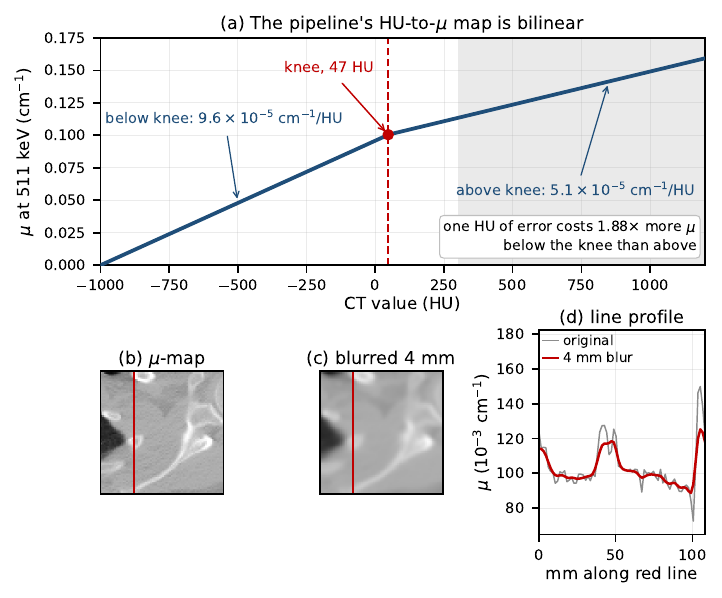}
\caption{
The two properties of the fixed reconstruction pipeline that shaped our loss.
\textbf{(a)} The Carney bilinear HU-to-$\mu$ map~\cite{carney2006hu2mu} the CT metric is computed in. 
Its knee at $47$\,HU makes one HU of error worth $1.88\times$ more $\mu$ below the knee than above, which is why we optimize in $\mu$ space rather than in HU; the shaded band marks the $>300$\,HU region our loss weights $\times 3$.
\textbf{(b, c)} A $110$\,mm window on the ribs and spine of training subject \texttt{sub-000}, before and after the $4$\,mm FWHM smoothing the pipeline applies before forward projection.
\textbf{(d)} Profile along the red line.
Smoothing removes $3.6$\,\% of the mean $\mu$ in this window and cuts the profile's peak-to-peak range from $77$ to $37\times10^{-3}$\,cm$^{-1}$, halving the narrow rib peaks.
Anything a model gets right at this scale is invisible to the three AC-PET metrics and visible only to the CT metric, which is the tension every design choice below addresses.
Both the conversion and the blur are computed with the organizers' own code.}
\label{fig:pipeline}
\end{figure}

Two properties of the fixed pipeline shaped our design, and Figure~\ref{fig:pipeline} shows both.
First, the HU-to-$\mu$ map is bilinear with a knee near $47$\,HU, and the segment below the knee is $1.88$ times steeper than the one above ($9.6\times10^{-5}$ against $5.1\times10^{-5}$\,cm$^{-1}$ per HU), so the same HU error costs almost twice as much $\mu$ in soft tissue as it does in bone.
Second, the pipeline smooths the $\mu$-map with a $4$\,mm Full Width at Half Maximum (FWHM) Gaussian before forward projection, which flattens the rib and vertebral peaks that carry most of a thorax slice's fine structure.
Pseudo-CT detail finer than about $4$\,mm is therefore invisible to the three AC-PET metrics, while the CT metric sees it directly.

\section{Method}

We first describe the network, which is shared by every model we report, and then the five loss terms.
Only two things ever change between our models: how many input channels the network reads, and which of the loss terms are switched on.

\subsection{Network}

\begin{algorithm}[t]
\caption{
\textbf{MuNet.} \texttt{Res} is a residual convolution block, \texttt{(+)} is channel concatenation, and \texttt{AttnGate} is an attention gate on the skip connection~\cite{oktay2018attnunet}.
The two auxiliary heads on lines 10 and 11 predict the CT at half and quarter resolution during training and are discarded at inference.
Only the input width $C$ on line 1 differs between our two variants.
The patch shape on line 1 is the $208^3$ configuration; $\mathcal{L}_{\text{LOR}}$ (Subsection~\ref{subsec:l_or}) replaces it with a $448\times448\times48$ slab, which changes no layer.
}
\label{fig:arch}
\begin{verbatim}
 1  x : (C,208,208,208)  C = 2 : NAC-PET, topogram      -> MuNet_2
 2                       C = 4 : + raw in/out MRI       -> MuNet_4
 3  s1 = Res(x,   32); e1 = Down(s1)
 4  s2 = Res(e1,  64); e2 = Down(s2)
 5  s3 = Res(e2, 128); e3 = Down(s3)
 6  s4 = Res(e3, 256); e4 = Down(s4)
 7  b  = Res(e4, 512)
 8  b  = b + SelfAttn(b, heads=8); b = Dropout(b, 0.2)
 9  d4 = Res(Up(b ) (+) AttnGate(s4, b ))
10  d3 = Res(Up(d4) (+) AttnGate(s3, d4)) -> aux_out2 (w=0.25) --.
11  d2 = Res(Up(d3) (+) AttnGate(s2, d3)) -> aux_out1 (w=0.50) -.|
12  d1 = Res(Up(d2) (+) AttnGate(s1, d2))                       ||
13  ct = Conv1x1x1(d1) in [0,1] -> HU = 3000*ct - 1000 <--------''
\end{verbatim}
\end{algorithm}

The organizers' baseline is a patch-based residual 3D U-Net~\cite{ronneberger2015unet} with dropout at the bottleneck, reading NAC-PET as its only input.
We keep that backbone and make three additions, shown in Listing~\ref{fig:arch}.
Attention gates on the skip connections let the decoder suppress the parts of each skip it does not need~\cite{oktay2018attnunet}.
One self-attention block sits at the bottleneck.
Two auxiliary heads predict the CT at half and quarter resolution, each carrying its own copy of the loss at a reduced weight, so the middle of the decoder receives a training signal directly rather than only through the layers above it.
We call the result \textbf{MuNet}.

Two input configurations are used: 
$\text{MuNet}_{2}$ reads NAC-PET and the topogram, whereas $\text{MuNet}_{4}$ extends this input with the raw in- and out-phase DIXON volumes as two additional channels.
The MRI is supplied on the CT grid but aligned only by a rigid translation, so roughly $40$\,mm of deformable misalignment remains, and $\text{MuNet}_{4}$ receives it uncorrected.
This is deliberate.
Registering the MRI to the ground-truth CT with ANTs SyN~\cite{avants2008syn} is worth $7.8$\,\% of CT $\mu$-MAE in a matched pair ($0.006476$ to $0.005970$), but it cannot be deployed, because validation and test subjects have no CT to register to.
Registering to a predicted guide CT can be deployed, yet then the operator differs between training and test, and it captured at most a sixth of that gain.
A learned registration network~\cite{balakrishnan2019voxelmorph} keeps the two operators identical but reached only $13.6$\,\% of oracle registration quality and captured none of the gain.
Feeding the MRI in unregistered avoids the problem rather than solving it: the residual misalignment is small relative to the bottleneck's receptive field, so the network can absorb it internally instead of depending on an operator we cannot keep consistent.

\subsection{$\mathcal{L}_{\mu}$: a region-weighted loss in the metric's own space}

Our single most effective change was to stop measuring the error in normalized HU and measure it where the CT metric measures it.
The metric converts HU to a linear attenuation coefficient $\mu$ with the Carney bilinear map before scoring, and that map is piecewise linear with a knee at $47$\,HU, so one HU of error is worth $1.88$ times more $\mu$ below the knee than above it (Figure~\ref{fig:pipeline}a).
An $L_1$ loss in HU therefore misprices its own errors relative to the metric, while an $L_1$ loss in $\mu$ prices them correctly.

The organizers supply a body mask per subject, and we score the loss only inside it.
It is the body minus the face and the scanner bed, because the reconstruction pipeline replaces those two regions with ground truth before it projects, so nothing a model predicts there can affect any metric.
Let $v$ index the voxels of a training patch, $x_v$ the network output in $[0,1]$, $y_v$ the target, $m_v$ that body mask, and $\mu(\cdot)$ the Carney map, which is differentiable.
We minimize
\begin{equation}
\mathcal{L}_{\mu} =
\frac{\sum_{v} w_v m_v \left| \mu(x_v) - \mu(y_v) \right|}{\sum_v w_v m_v},
\quad
w_v = \max\!\big(1, \lambda^{\text{bone}}_v, \lambda^{\text{brain}}_v, \lambda^{\text{organ}}_v\big).
\label{eq:mul1}
\end{equation}
The three region factors up-weight the places the four scored metrics are sensitive to: $\lambda^{\text{bone}}_v = 3$ where the ground-truth CT exceeds $300$\,HU, $\lambda^{\text{brain}}_v = 3$ inside the brain, and $\lambda^{\text{organ}}_v = 2$ inside the four organs the challenge's organ-bias metric is computed over, namely spleen, liver, pancreas and heart.
We did not choose those four; the metric does, and we simply weight what it measures.
Taking the maximum rather than the product keeps the combined weight bounded by $3$, so no voxel can dominate the batch.
The regions come from segmentations of the ground-truth CT and are used only to form the weights during training, so the deployed model still reads nothing but the released features.

These weights are also what makes training from random initialization possible at all.
Under a plain $L_1$ objective our from-scratch runs produced pseudo-CTs with essentially no voxels above $300$\,HU: the network never learned bone, which ruins the brain-outlier score because that metric is driven by the skull.

\subsection{$\mathcal{L}_{\text{grad}}$: matching edge strength}

$\mathcal{L}_{\mu}$ is a per-voxel loss and is therefore indifferent to whether an edge is sharp or smeared, as long as the average intensity is right.
$\mathcal{L}_{\text{grad}}$ penalizes the absolute difference between the gradient magnitudes of the predicted and target CT, which asks for the same edge strength in the same places, and mainly affects the soft-tissue boundaries where our error concentrates.

\subsection{$\mathcal{L}_{\text{AnPer}}$: an Anatomical Perception loss}

A pseudo-CT can have a low voxel-wise error and still be anatomically implausible in ways the AC-PET metrics punish.
$\mathcal{L}_{\text{AnPer}}$ addresses this by passing the predicted CT and the ground-truth CT through a frozen TotalSegmentator~\cite{wasserthal2023totalsegmentator} network and matching intermediate decoder features, in the spirit of perceptual losses~\cite{johnson2016perceptual}.
Because the segmenter was trained to recognise organs, agreeing in its feature space means agreeing about anatomy rather than about intensities.
It runs in training only, and it moved the AC-PET metrics more than anything else we tried: in a matched pair differing by nothing else it improved organ bias by $5.8$\,\% ($3.34$ to $3.14$) and the brain-outlier score by $27.4$\,\% ($0.0261$ to $0.0189$).

\subsection{$\mathcal{L}_{\text{NGF}}$: aligning boundaries with the MRI}

Decomposing $\text{MuNet}_2$'s error shows that soft-tissue boundaries carry $52.6$\,\% of its CT $\mu$-MAE ($0.0031$ of $0.0058$) against $20.9$\,\% ($0.0012$) for soft-tissue interiors, so the boundaries are where the remaining error lives.
DIXON MRI has strong soft-tissue contrast and almost no bone signal, which makes it a good boundary prior and a poor intensity target.
$\mathcal{L}_{\text{NGF}}$ therefore compares the two images' gradient directions while ignoring their magnitudes, using a Normalized Gradient Field (NGF)~\cite{haber2006ngf}: it penalizes the cross product of the unit gradient fields of the predicted CT and the mean of the two MRI phases, which is zero when the two images have edges in the same places regardless of how differently they are scaled.
We gate it to soft tissue, where the MRI is informative.
We keep the name from the original formulation because it is standard in registration, where the same construction is used to align images of different modalities.
The MRI enters through this term alone, so a model trained with it still reads only NAC-PET and the topogram at inference.

\subsection{$\mathcal{L}_{\text{LOR}}$: a loss along \textbf{L}ines \textbf{O}f \textbf{R}esponse}
\label{subsec:l_or}

Every term so far is local, and three of the four metrics are not.
The attenuation along a line of response is $\exp(-\!\int\!\mu\,ds)$, so the reconstructed activity in an organ depends on $\mu$ integrated along lines that cross the whole body, not on $\mu$ inside that organ.
Making an organ's own $\mu$ accurate therefore does not make its reconstructed activity accurate, which is why simply raising $\lambda^{\text{organ}}$ failed to improve organ bias.

$\mathcal{L}_{\text{LOR}}$ penalizes $\left|\int (\mu_{\text{pred}} - \mu_{\text{gt}})\,ds\right|$ along transaxial lines at four angles, computed as a Radon transform per axial slice, separately for lines crossing the whole body and for lines crossing each scored organ.
Two details matter.
We smooth the $\mu$ difference at $4$\,mm FWHM before integrating, matching the blur the pipeline applies before its own forward projection (Figure~\ref{fig:pipeline}b--d), so the term only ever sees the band of spatial frequencies the AC-PET metrics can see.
And a line integral is meaningless unless the line spans the body, so this term forces the training patch from a $208^3$ cube to a $448\times448\times48$ slab, which gives $681$\,mm of transaxial support at $1.07$ times the voxel count, leaving the number of patches per batch unchanged.

\subsection{Fusion of finished models}

The container writes a single pseudo-CT and the organizers run the reconstruction themselves, so we cannot submit one model's CT with another model's PET: any combination of models has to collapse into one volume before it is scored.
We therefore average finished models voxel-wise in HU, $\mathrm{CT}_{\text{fused}} = \sum_i \alpha_i \mathrm{CT}_i$ with positive weights summing to one, fixed in the image.
The members train independently, so their errors are partly independent and averaging cancels some of them~\cite{lakshminarayanan2017ensembles}.
We chose the pair and its weights by analysing the members' measured metrics, then confirmed the choice with a full reconstruction; we also tried three-member blends, which scored no better than the best pair and cost CT accuracy, so we report only the pair.

\section{Results}

All models train on all 75 released subjects with Adam~\cite{kingma2015adam} and a cosine schedule~\cite{loshchilov2017sgdr}.
Patches are $208^3$, or $448\times448\times48$ where $\mathcal{L}_{\text{LOR}}$ is active, two patches per optimizer step in both cases, which fills the $48$\,GB of one RTX A6000.
A model takes $18$ to $60$ hours to train depending on the schedule.
Table~\ref{tab:progression} follows the order we built things in, and each row differs from the one above it by exactly one change.

Rows 1 to 2 are the loss alone: same inputs, same backbone, no MRI, and CT $\mu$-MAE falls from $0.006610$ to $0.005803$.
That is the largest single step in the table, and it is why we say the loss mattered more than the architecture.
Row 3 adds $\mathcal{L}_{\text{NGF}}$ and row 5 instead adds the MRI as input channels, and the two do opposite things.
Row 5 gives the best CT $\mu$-MAE of any single model, $0.005679$, which is the evidence that the unregistered MRI carries usable information.
Row 3 gives up CT ($0.005972$) and buys the AC-PET metrics instead.
The $4$\,mm blur is why: a term that moves fine structure changes the CT metric directly, because that metric sees the pseudo-CT at full resolution, but most of its effect is smoothed away before the projection the other three metrics depend on.

\begin{table}[t]
\caption{
Validation-set progression, at the precision the leaderboard reports.
Each row adds the change in the \emph{Configuration} column to the row above it within its block.
Best per column in \textbf{bold}.
Row 9 is the model we submit.
Row 7's three AC-PET entries are marked \texttt{--} because the challenge's validation phase closed before we scored that model on its own, and those metrics can only be produced by the organizers' service.
Its CT $\mu$-MAE is measured, and its effect on all four metrics is visible in row 9.
}
\label{tab:progression}
\centering
{\fontsize{8}{10}\selectfont
\setlength{\tabcolsep}{3.5pt}
\begin{tabular}{@{}rlp{3.7cm}cccc@{}}
\toprule
\# & Model & Configuration & CT $\mu$-MAE\,$\downarrow$ & SUV MAE\,$\downarrow$ & Organ\,$\downarrow$ & Brain\,$\downarrow$ \\
\midrule
1 & Baseline & plain $L_1$, NAC-PET only & 0.006610 & 0.0586 & 4.56 & 0.0541 \\
\midrule
2 & \multirow{3}{*}{$\text{MuNet}_{2}$}
    & $\mathcal{L}_{\mu}+\mathcal{L}_{\text{grad}}+\mathcal{L}_{\text{AnPer}}$ & 0.005803 & 0.0377 & 2.74 & 0.0148 \\
3 & & \;$+\ \mathcal{L}_{\text{NGF}}$ & 0.005972 & 0.0352 & 2.58 & 0.0158 \\
4 & & \;$+\ \mathcal{L}_{\text{LOR}}$ & 0.005958 & 0.0345 & 2.45 & 0.0168 \\
\midrule
5 & \multirow{3}{*}{$\text{MuNet}_{4}$}
    & $\mathcal{L}_{\mu}+\mathcal{L}_{\text{grad}}+\mathcal{L}_{\text{AnPer}}$ & 0.005679 & 0.0383 & 2.80 & 0.0168 \\
6 & & \;$+\ \mathcal{L}_{\text{LOR}}$ & 0.005703 & 0.0339 & 2.38 & \textbf{0.0078} \\
7 & & \;$+\ 100$ epochs at a fifth of the learning rate & 0.005710 & -- & -- & -- \\
\midrule
8 & \multirow{2}{*}{Fusion}
    & $0.3\times$ row 4 $+\ 0.7\times$ row 6 & \textbf{0.005627} & 0.0331 & 2.27 & 0.0091 \\
9 & & $0.3\times$ row 4 $+\ 0.7\times$ row 7 & 0.005633 & \textbf{0.0326} & \textbf{2.23} & 0.0087 \\
\bottomrule
\end{tabular}}
\end{table}

Rows 4 and 6 add $\mathcal{L}_{\text{LOR}}$ to each of them, and both shift the same way.
Row 6 against row 5 is the clearest case: $0.42$\,\% worse on CT, and $11.4$\,\%, $15.0$\,\% and $53.9$\,\% better on SUV MAE, organ bias and brain outlier.
It is also the only term we found that improved organ bias at all.
Row 8 averages rows 4 and 6.

Row 7 refines row 6 for $100$ further epochs at a fifth of the learning rate, and row 9 replaces row 6 with it inside the same blend.
Row 9 is the model we submit.
The refinement left the training loss where it started ($0.0133$ against $0.0132$), and row 9 still improves on row 8 by $1.3$\,\% on SUV MAE, $1.9$\,\% on organ bias and $4.9$\,\% on brain outlier, for $0.11$\,\% more CT $\mu$-MAE.
We report both blends because the training loss gave no sign of that gain, so a run gated on training loss alone would have thrown the refinement away.

The table has one gap: we never trained $\text{MuNet}_{4}$ with $\mathcal{L}_{\text{NGF}}$.
The two are partly redundant by construction, because $\mathcal{L}_{\text{NGF}}$ exists to inject MRI information into a network that cannot see the MRI, whereas $\text{MuNet}_{4}$ reads it directly.
We would expect less from the combination for that reason, but we have not measured it, and we would rather record the gap than argue it away.

\subsection{Why the fusion gain depends on the pair}

Row 8 beats both of its members on CT $\mu$-MAE, at $0.005627$ against $0.005703$ and $0.005958$, which is $2.6$\,\% below the value linear interpolation between them would give.
The effect on the brain-outlier score is larger still, and we think the shape of that metric is the reason: it is an area under a curve of per-subject error fractions, so it is a tail statistic, and averaging two partly independent predictors suppresses tails much more than it shifts means.

The size of that gain therefore depends on which two models are averaged rather than on how good either is alone, and rows 4 to 6 show both sides of it.
Row 4 has a brain-outlier score of $0.0168$, the same value as row 5, and averaging those two gains almost nothing on brain because they fail on the same subjects: a blend we built that way lost three leaderboard places on brain while gaining one on CT, even though it was better on the other three metrics.
Row 6 scores $0.0078$ on the same metric, and averaging row 4 with row 6 instead gives row 8.
The same model is a poor partner in one pair and half of our best result in the other, so we chose the pair from the members' measured metrics and confirmed it with a full reconstruction rather than trusting any single model's numbers.

\subsection{Leaderboard}

On the public validation leaderboard of 16 August 2026, the submitted fusion ranks \textbf{first overall} with a mean rank of $3.50$ across 26 submissions, with per-metric ranks 7th on CT $\mu$-MAE, 2nd on SUV MAE, 3rd on organ bias and 2nd on brain outlier.
The margin is thin and the shape of it is worth stating: the second-placed entry, at $3.88$, beats us on three of the four metrics and loses only on CT $\mu$-MAE, where we are $3.3$\,\% ahead.

\bibliographystyle{splncs04}
\bibliography{ref}

\end{document}